\documentclass{article}

    \PassOptionsToPackage{numbers, compress}{natbib}
    \usepackage[final]{neurips_2024_ml4ps}

\usepackage[utf8]{inputenc} % allow utf-8 input
\usepackage[T1]{fontenc}    % use 8-bit T1 fonts
\usepackage{hyperref}       % hyperlinks
\usepackage{url}            % simple URL typesetting
\usepackage{booktabs}       % professional-quality tables
\usepackage{amsfonts}       % blackboard math symbols
\usepackage{nicefrac}       % compact symbols for 1/2, etc.
\usepackage{microtype}      % microtypography
\usepackage{xcolor}         % colors
\usepackage[pdftex]{graphicx}
\usepackage{amsmath}
\usepackage{todonotes}
\usepackage{float}

\def\code#1{\texttt{#1}}

\title{Harnessing Machine Learning for Single-Shot Measurement of Free Electron Laser Pulse Power}

\author{%
  Till Korten \\
  Helmholtz AI Team Matter\\
  Helmholtz-Zentrum Dresden-Rossendorf HZDR\\
  01328 Dresden Germany \\
  \And
  Vladimir Rybnikov \\
  Deutsches Elektronen-Synchrotron DESY \\
  22607 Hamburg, Germany \\
  \And
  Mathias Vogt \\
  Deutsches Elektronen-Synchrotron DESY \\
  22607 Hamburg, Germany \\
  \And
  Juliane Roensch-Schulenburg \\
  Deutsches Elektronen-Synchrotron DESY \\
  22607 Hamburg, Germany \\
  \And
  Peter Steinbach \\
  Helmholtz AI Team Matter\\
  Helmholtz-Zentrum Dresden-Rossendorf HZDR\\
  01328 Dresden Germany \\
  \And
  Najmeh Mirian \\
  Institute of Radiation Physics\\
  Helmholtz-Zentrum Dresden-Rossendorf HZDR\\
  01328 Dresden Germany \\
  \texttt{n.mirian@hzdr.de} \\
}

\begin{document}

\maketitle

\begin{abstract}
Electron beam accelerators are essential in many scientific and technological fields. Their operation relies heavily on the stability and precision of the electron beam. Traditional diagnostic techniques encounter difficulties in addressing the complex and dynamic nature of electron beams. Particularly in the context of free-electron lasers (FELs), it is fundamentally impossible to measure the lasing-on and lasing-off electron power profiles for a single electron bunch. This is a crucial hurdle in the exact reconstruction of the photon pulse profile. To overcome this hurdle, we developed a machine learning model
%\footnote{See section \ref{sec:availability} for code and data availability.} 
that predicts the temporal power profile of the electron bunch in the lasing-off regime using machine parameters that can be obtained when lasing is on. The model was statistically validated and showed superior predictions compared to the state-of-the-art batch calibrations. The work we present here is a critical element for a virtual pulse reconstruction diagnostic (VPRD) tool designed to reconstruct the power profile of individual photon pulses without requiring repeated measurements in the lasing-off regime. This promises to significantly enhance the diagnostic capabilities in FELs at large.
\end{abstract}

\section{Introduction}

Electron beam accelerators are fundamental to numerous scientific and technological fields, encompassing medical therapies, material science, and particle physics research. Their effective operation depends heavily on the stability and precision of the electron beam, which in turn relies on advanced diagnostic technologies to monitor and maintain beam quality. Traditional diagnostic techniques, while valuable, often encounter difficulties in addressing the complex and dynamic nature of electron beams. This is where machine learning (ML) has emerged as a transformative technology, offering enhanced diagnostic capabilities. By improving precision, adaptability, and real-time analysis, ML-driven diagnostics are reshaping how electron beams are monitored and optimized\cite{ratner2020,Kaiser2024, Kaiser2024_2, Fujita2021, Florian2020}.

To achieve the required precision in experiments of free-electron lasers (FELs), various diagnostic methods have been developed. One such method is the transverse reconstruction of the electron beam and X-ray algorithm \cite{TREX,SLAC}.
This approach aims to reconstruct the temporal power profile $I(t)$ of the beam. It does so by analyzing the product of the measured current profile and the difference between the energy spread or mean energy profiles of lasing-on and lasing-off shots using a transverse deflecting cavity (TDS) (see Figure 2 in \cite{SLAC}). However, this approach is constrained by the finite resolution of the TDS \cite{TDS}. In addition, it is impossible to measure the lasing-on\footnote{a high intensity photon beam is produced by the FEL} and lasing-off\footnote{no photon beam is produced by the FEL} electron phase spaces for a single shot \cite{ Hanuka2021}. 
This means that during an experiment - where lasing must be on - the lasing-off electron phase space is inaccessible to experimental measurement.
Currently, this limitation is worked around by performing batch calibrations: several hundred temporal power profiles of electron bunches are recorded without lasing (lasing-off regime), followed by several hundred shots with lasing (lasing-on regime). The power profiles of these shots are then averaged for each context and subtracted from each other. This produces a mean difference for lasing-on versus lasing-off power profiles. This allows us to calculate the mean temporal power profile $I(t)$ of the photon pulse \cite{TREX}. However, the state of the art approach cannot account for differences between individual pulses.

\begin{figure}[htb]
    \centering
    \includegraphics[width=0.9\linewidth]{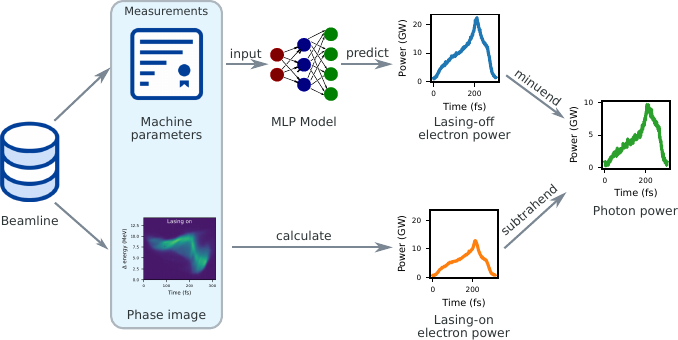}
    \caption{Process workflow. During beamline operation, we obtain measured \emph{"machine parameters}" and the longitudinal phase space (\emph{"phase image}") for each photon pulse. We use a multi-layer perceptron machine learning model (\emph{"MLP model}") to predict the temporal power profile of the electron bunch if it would be obtained without lasing (\emph{"lasing-off electron power}"). From the phase image, we calculate the temporal power profile of the electron bunch with lasing (\emph{"lasing-on electron power"}). Thus, we can estimate the temporal power profile for each individual photon pulse (\emph{"Photon power"}) by subtracting the measured lasing-on electron power from the predicted lasing-off electron power. The top row (machine parameters $\rightarrow$ MLP model $\rightarrow$ lasing-off electron power) is the part of the process that we solved in this paper.}
    \label{fig:scheme}
\end{figure}

This paper introduces an important step towards Virtual Pulse Reconstruction Diagnostic (VPRD), a tool designed to address the limitations of traditional diagnostic methods, particularly in the context of FELs. VPRD will leverage machine learning algorithms to reconstruct the longitudinal phase space of individual pulses of the electron beam without the need for repeated measurements in the lasing-off regime, offering a non-invasive and efficient method for characterizing free electron laser radiation pulses. Towards this end, VPRD will use the following workflow (Figure \ref{fig:scheme}):
For each electron bunch, the beam monitoring system records $22$ machine parameters (Table \ref{tab:machine_parameters}) as well as a longitudinal phase space image in the lasing-on regime. The machine par ameters are used as input to predict the shape that the temporal power profile of the electron bunch would have in the lasing-off regime. The temporal power profile in the lasing-on regime is measured directly from the longitudinal phase space. Thus, we will be able to reconstruct the temporal power profile for each individual photon pulse.

In this paper, we focus on the machine-learning part of the approach motivated above. Our contributions are as follows: We developed a multi-layer perceptron (MLP) machine learning model (section \ref{sec:training}) that was able to predict the temporal power profile of the electron bunch in the lasing-off regime (section \ref{sec:results}). Towards this end, we collected the machine parameters as well as the longitudinal phase space for 2826 electron bunches in the lasing-off regime as training data. We statistically validate predictions of our approach by comparing them to state of the art \cite{TREX, SLAC} on a test set (section \ref{sec:results}). Finally, we discuss the impact and limitations of our results (section \ref{sec:discussion}).

\section{Methodology}
\label{sec:methodology}

\subsection{Model training and validation}
\label{sec:training}
The training dataset comprised $2826$ samples. Each sample consisted of $22$ measured machine parameters as input and a lasing-off electron power profile as label. The electron power profiles were calculated from a measured phase image as described in appendix \ref{sec:power-calculation}.

We used a MLP model with $22$ input nodes (see Table \ref{tab:machine_parameters} for details on the input parameters), a single hidden layer with $294$ nodes and an output layer with $567$ nodes (the width of the electron temporal power profiles in the training data). The MLP model was set up to perform a regression task predicting a signal of size $567$ given an input sample of size $22$.
The $2826$ samples of our dataset were split into training, validation and test sets with $2261$, $283$ and $282$ samples respectively. For this, we used \texttt{torch.utils.data.random\_split} with a random seed of $42$ to obtain reproducible results. 

We used the mean squared error loss function \code{torch.nn.MSELoss()}. As an alternative, we explored an adapted loss function that penalizes regression to the mean:

\begin{equation}
    L = \sum_{i=1}^{D}(x_i-y_i)^2 - \alpha\sum_{i=1}^{D}(x_i-\hat{y})^2
\label{eqn:loss}
\end{equation}
where $\hat{y}$ is the mean of the labels of the entire training dataset, $\alpha$ is a penalty factor, $x_i$ are the predictions and $y_i$ are the respective labels. Comparing the mean squared error on the test set showed that the alternative loss function led to models performing worse for $\alpha > 0.05$.

Hyperparameters were optimized with optuna \cite{akiba2019optuna} using 200 trials. After optimization, a model with one hidden layer was trained in Pytorch \cite{Ansel_2024} using the Adam optimizer \cite{Adam_2017},a dropout fraction of 0.45 on the hidden layer, an initial learning rate of 0.005, a learning rate scheduler with a factor 0f 0.05 and a patience of 238. Training was stopped using the \code{EarlyStopping} callback from Pytorch Lightning \cite{Falcon_PyTorch_Lightning_2019} with a patience of 1225.  To improve GPU utilization, we used single-batch training. Thus, training the model was conducted on the entire training dataset and validating on the entire validation dataset at once for each training step. Training and validation losses converged well with no indication of overfitting (Figure \ref{fig:ml_results}a). Note that the validation loss is lower than the training loss (clearly visible in Figure \ref{fig:ml_results}a), because of the dropout used during training. Training the model took approximately 1 min on a laptop GPU (Apple M3 Pro). Data preprocessing took approximately 2.5 min. During development and testing, we trained approximately 300 models for this project, using approximately 5 GPU hours in total for training and on the order of 10 GPU hours for data preprocessing.

\subsection{Data and code availability}
\label{sec:availability}
%PS: Put the data and code section in the appendix and add a footnote in the main text to it!

The code is available on GitHub \cite{code}.

The data is available on RoDaRe \cite{data}.

\section{Results}
\label{sec:results}

\begin{figure}[!htb]
    \centering
    \includegraphics[width=0.9\linewidth]{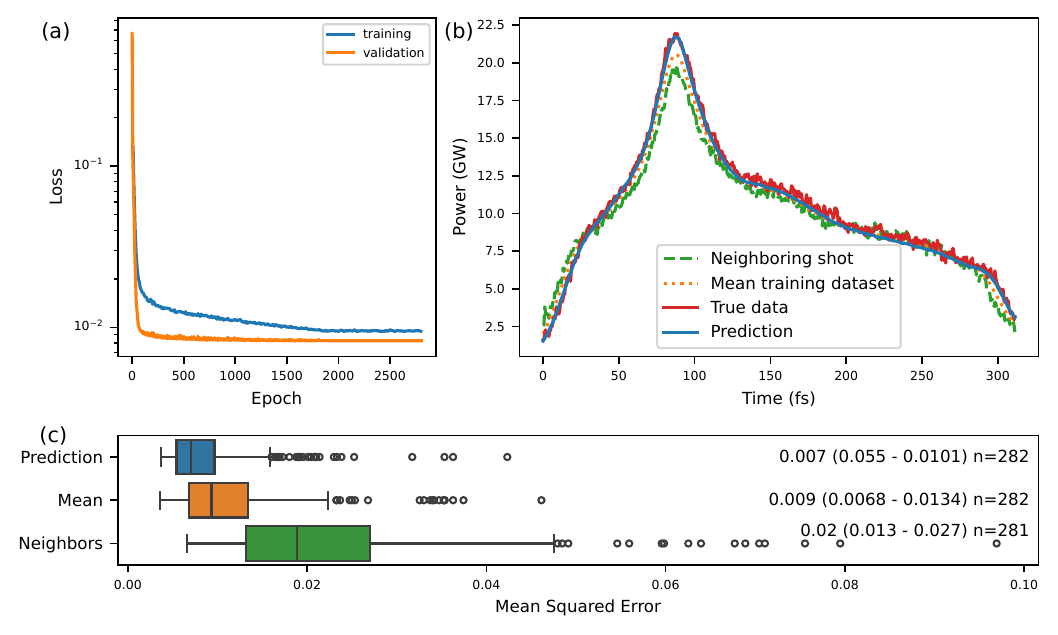}
    \caption{ MLP model training performance. (a) Training and validation loss. The validation loss is lower than the training loss, because we use a dropout of 0.43 during training. (b) Predictions for individual shots (blue line) matched the actual measurements (red line) better than measurements from previous shots (dashed green line) and better than the mean of all measurements in the training data (dotted orange line). (c) Boxplots of all mean squared errors in the test dataset. (Prediction) Mean squared error between the predictions and the measurements in the test dataset. (Mean) Mean squared error between the measurements in the test dataset and the mean of all measurements in the training dataset. (Neighbors) Mean squared error between adjacent measurements in the test dataset. The numbers at the right of (c) represent the median (interquartile range) of the errors as well as the number of observations $n$.}
    \label{fig:ml_results}
\end{figure}

Predicted temporal power profiles matched the measured profiles very well (Figure \ref{fig:ml_results}b blue line vs. red line). We did not observe regression to the mean of the training data set (Figure \ref{fig:ml_results}b orange dotted line). Neighboring shots have been used as labels to train machine learning models to predict longitudinal phase space data in the past \cite{Zhu_2022}. Therefore, we compared the measured profile to the neighboring measurement (Figure \ref{fig:ml_results}b green dashed line vs. red line). However, neighboring shots fit the measurement even worse than the mean of all shots. 

These individual observations exemplified in (Figure \ref{fig:ml_results}b) are confirmed by plotting the mean squared errors for all samples in the test dataset (Figure \ref{fig:ml_results}c). The predictions of the MLP model have the lowest mean squared error compared to the measurements in the test dataset (Prediction; $0.007 (0.055 - 0.0101)$ (median and interquartile range) $n=282$)). This is better than the mean squared error between the mean of the entire training dataset and the individual measurements in the test dataset (Mean: $0.009 (0.0068 - 0.0134)$ $n=282$). The highest mean squared error was observed for neighboring measurements in the test dataset ($0.02 (0.013 - 0.027)$ $n=281$).
%%%%%%%%%%%%%%%%%%%%%%%%%%
%%%%%%%%%%%%%%%

The prediction error was statistically significantly lower than both the mean and the neighbor errors ($p<0.01$) as determined by a Wilcoxon signed-rank test \cite{Wilcoxon_1945} followed by a Bonferroni correction\cite{Bonferroni_1936} for multiple comparisons (making no assumptions about the data distribution). This analysis underpins the superiority of the model proposed in this paper.
\section{Discussion}
\label{sec:discussion}
Our model delivers superior predictions for the electron temporal power profile than the state-of-the-art batch calibrations. The observation that neighboring measurements are not a good predictor is significant, because neighboring shots have been used as labels to train machine learning models to predict longitudinal phase space data in the past \cite{Zhu_2022}.

\subsection{Limitations}
\label{sec:limitations}
In order to be useful for the intended use, i.e. predicting lasing-off electron temporal power profiles while the beamline is operating in lasing-on mode, we are assuming that the machine parameters we used to train the model are independent of whether lasing is on or off. We believe this assumption is valid, because all parameters used for training are measured before the undulator magnet that induces lasing. However, this assumption needs to be tested in future experiments.

The model is  in principle computationally efficient enough to allow predictions in the high kHz range (predictions take $16.1 \pm 0.80$ µs per shot on a laptop computer). Nevertheless, it is currently not realistic to integrate predictions into a live monitoring workflow due to technical limitations. However this may be possible in the future if the model is implemented on a field programmable array that is integrated into the existing beamline monitoring system. For most applications, it is currently good enough to determine the photon pulse shape after the experiment is done.

Currently, we were using a relatively small set of $2826$ samples. While this is not a lot compared to other deep learning projects, it was enough to demonstrate superior predictions compared to the state of the art. In the future, we will collect larger datasets and explore whether this will improve the predictions further. We will also explore whether it is possible to pre-train the model on artificial data.

\subsection{Conclusions}

In this paper, we have demonstrated a machine learning approach for predicting the electron beam temporal power profile in the lasing-off regime from machine parameters collected in accelerator section and the beam line. During regular beam line operation, this lasing-off data is inaccessible. Unlike conventional methods, our machine-learning-driven approach overcomes the limitations of single-shot measurements, as well as batch calibrations, offering a non-invasive and efficient method for characterizing free electron laser radiation pulses. In the future we will test the machine learning model during regular operation aiming to enable single-shot measurements of the X-ray pulse structure in real-time. This can open new avenues for scientific discovery for a wide range of FEL experiments. In addition, aspects of trustworthy use of machine learning need to be studied for this application of the model, such as interpretability, robustness and uncertainty quantification.

\begin{ack}
%Use unnumbered first level headings for the acknowledgments. All acknowledgments go at the end of the paper before the list of references. Moreover, you are required to declare funding (financial activities supporting the submitted work) and competing interests (related financial activities outside the submitted work).More information about this disclosure can be found at: \url{https://neurips.cc/Conferences/2024/PaperInformation/FundingDisclosure}.
%Do {\bf not} include this section in the anonymized submission, only in the final paper. You can use the \texttt{ack} environment provided in the style file to automatically hide this section in the anonymized submission.

\subsection*{Acknowledgments}
The authors like to express the gratitude to the late Siegfried Schreiber, former head of the FLASH facility, for his support and encouragement in the early stages of this project.

We also thank members of the FLASH operation team for providing help and conditions to carry out the data collection.

We thank the entire Team of Helmholtz AI Matter for invaluable discussions and a great working atmosphere.

\subsection*{Author contributions}
All authors contributed to writing the manuscript. In addition:
\begin{description}
\item[TK]{conducted data processing, trained the machine learning model, and prepared the experimental figures.}
\item[VR] {provided the data acquisition system in the FLASH control system and supported data collection.}
\item[MV and JSR] {prepared the FLASH machine, supported the project, and contributed to coordinating data collection activities}
%\item{JRS} prepared the FLASH machine and support the project.
\item[PS]{coordinated and providing guidance on the leveraging machine learning techniques and analyzing the data}
\item[NM]{conceived the original idea and concept, led the project, collected and pre-processed the experimental data} %supported TK with understanding and processing the data and provided the original idea and concept for the paper 
%\todo{@Naji please complete author contributions for yourself and the other %authors from DESY}}
\end{description}

\subsection*{Funding}
The work of TK and PS was funded by Helmholtz Incubator Platform Helmholtz AI. %\todo{@Naji please add your funding sources} I don't have it :-)
\end{ack}

% \section*{References}

\bibliography{references}

\begin{thebibliography}{27}
\providecommand{\natexlab}[1]{#1}
\providecommand{\url}[1]{\texttt{#1}}
\expandafter\ifx\csname urlstyle\endcsname\relax
  \providecommand{\doi}[1]{doi: #1}\else
  \providecommand{\doi}{doi: \begingroup \urlstyle{rm}\Url}\fi

\bibitem[Ratner(2020)]{ratner2020}
Daniel Ratner.
\newblock Introduction to {Machine} {Learning} for {Accelerator} {Physics},
  June 2020.
\newblock URL \url{http://arxiv.org/abs/2006.09913}.
\newblock arXiv:2006.09913.

\bibitem[Kaiser et~al.(2024{\natexlab{a}})Kaiser, Xu, Eichler,
  Santamaria~Garcia, Stein, Br{\"u}ndermann, Kuropka, Dinter, Mayet, Vinatier,
  Burkart, and Schlarb]{Kaiser2024}
Jan Kaiser, Chenran Xu, Annika Eichler, Andrea Santamaria~Garcia, Oliver Stein,
  Erik Br{\"u}ndermann, Willi Kuropka, Hannes Dinter, Frank Mayet, Thomas
  Vinatier, Florian Burkart, and Holger Schlarb.
\newblock Reinforcement learning-trained optimisers and bayesian optimisation
  for online particle accelerator tuning.
\newblock \emph{Scientific Reports}, 14\penalty0 (1):\penalty0 15733, Jul
  2024{\natexlab{a}}.
\newblock ISSN 2045-2322.
\newblock \doi{10.1038/s41598-024-66263-y}.
\newblock URL \url{https://doi.org/10.1038/s41598-024-66263-y}.

\bibitem[Kaiser et~al.(2024{\natexlab{b}})Kaiser, Xu, Eichler, and
  Santamaria~Garcia]{Kaiser2024_2}
Jan Kaiser, Chenran Xu, Annika Eichler, and Andrea Santamaria~Garcia.
\newblock Bridging the gap between machine learning and particle accelerator
  physics with high-speed, differentiable simulations.
\newblock \emph{Phys. Rev. Accel. Beams}, 27:\penalty0 054601, May
  2024{\natexlab{b}}.
\newblock \doi{10.1103/PhysRevAccelBeams.27.054601}.
\newblock URL
  \url{https://link.aps.org/doi/10.1103/PhysRevAccelBeams.27.054601}.

\bibitem[Fujita(2021)]{Fujita2021}
Kazuhiro Fujita.
\newblock Physics-informed neural network method for space charge effect in
  particle accelerators.
\newblock \emph{IEEE Access}, 9:\penalty0 164017--164025, 2021.
\newblock \doi{10.1109/ACCESS.2021.3132942}.
\newblock URL \url{https://doi.org/10.1109/ACCESS.2021.3132942}.

\bibitem[Christie et~al.(2020)Christie, Lutman, Ding, Huang, Jhalani,
  Krzywinski, Maxwell, Ratner, R{\"o}nsch-Schulenburg, and Vogt]{Florian2020}
Florian Christie, Alberto~Andrea Lutman, Yuantao Ding, Zhirong Huang, Vatsal~A.
  Jhalani, Jacek Krzywinski, Timothy~J. Maxwell, Daniel Ratner, Juliane
  R{\"o}nsch-Schulenburg, and Mathias Vogt.
\newblock Temporal x-ray reconstruction using temporal and spectral
  measurements at lcls.
\newblock \emph{Scientific Reports}, 10\penalty0 (1):\penalty0 9799, 2020.
\newblock \doi{10.1038/s41598-020-66220-5}.
\newblock URL \url{https://doi.org/10.1038/s41598-020-66220-5}.

\bibitem[Ding et~al.(2011)Ding, Behrens, Emma, Frisch, Huang, Loos, Krejcik,
  and Wang]{TREX}
Yuantao Ding, Carsten Behrens, Paul Emma, Jérôme Frisch, Zhirong Huang,
  H.~Loos, P.~Krejcik, and M-H. Wang.
\newblock Femtosecond x-ray pulse temporal characterization in free-electron
  lasers using a transverse deflector.
\newblock \emph{Physical Review Special Topics - Accelerators and Beams},
  14\penalty0 (12):\penalty0 120701, December 2011.
\newblock \doi{10.1103/PhysRevSTAB.14.120701}.
\newblock URL \url{https://link.aps.org/doi/10.1103/PhysRevSTAB.14.120701}.
\newblock Publisher: American Physical Society.

\bibitem[Behrens et~al.(2014)Behrens, Decker, Ding, Dolgashev, Frisch, Huang,
  Krejcik, Loos, Lutman, Maxwell, Turner, Wang, Wang, Welch, and Wu]{SLAC}
C.~Behrens, F.-J. Decker, Yuantao Ding, V.~A. Dolgashev, J.~Frisch, Zhirong
  Huang, P.~Krejcik, H.~Loos, A.~Lutman, T.~J. Maxwell, J.~Turner, J.~Wang,
  M.-H. Wang, J.~Welch, and J.~Wu.
\newblock Few-femtosecond time-resolved measurements of x-ray free-electron
  lasers.
\newblock \emph{Nature Communications}, 5\penalty0 (1):\penalty0 3762, Apr
  2014.
\newblock ISSN 2041-1723.
\newblock \doi{10.1038/ncomms4762}.
\newblock URL \url{https://doi.org/10.1038/ncomms4762}.

\bibitem[Akre et~al.(2001)Akre, Bentson, Emma, and Krejcik]{TDS}
R.~Akre, L.~Bentson, P.~Emma, and P.~Krejcik.
\newblock A transverse rf deflecting structure for bunch length and phase space
  diagnostics.
\newblock In \emph{{PACS2001}. {Proceedings} of the 2001 {Particle}
  {Accelerator} {Conference} ({Cat}. {No}.{01CH37268})}, volume~3, pages
  2353--2355 vol.3, June 2001.
\newblock \doi{10.1109/PAC.2001.987379}.
\newblock URL \url{https://ieeexplore.ieee.org/document/987379}.

\bibitem[Hanuka et~al.(2021)Hanuka, Emma, Maxwell, Fisher, Jacobson, Hogan, and
  Huang]{Hanuka2021}
A.~Hanuka, C.~Emma, T.~Maxwell, A.~S. Fisher, B.~Jacobson, M.~J. Hogan, and
  Z.~Huang.
\newblock Accurate and confident prediction of electron beam longitudinal
  properties using spectral virtual diagnostics.
\newblock \emph{Scientific Reports}, 11\penalty0 (1):\penalty0 2945, Feb 2021.
\newblock ISSN 2045-2322.
\newblock \doi{10.1038/s41598-021-82473-0}.
\newblock URL \url{https://doi.org/10.1038/s41598-021-82473-0}.

\bibitem[Akiba et~al.(2019)Akiba, Sano, Yanase, Ohta, and
  Koyama]{akiba2019optuna}
Takuya Akiba, Shotaro Sano, Toshihiko Yanase, Takeru Ohta, and Masanori Koyama.
\newblock Optuna: {A} {Next}-generation {Hyperparameter} {Optimization}
  {Framework}.
\newblock In \emph{Proceedings of the 25th {ACM} {SIGKDD} {International}
  {Conference} on {Knowledge} {Discovery} \& {Data} {Mining}}, {KDD} '19, pages
  2623--2631, New York, NY, USA, July 2019. Association for Computing
  Machinery.
\newblock ISBN 978-1-4503-6201-6.
\newblock \doi{10.1145/3292500.3330701}.
\newblock URL \url{https://doi.org/10.1145/3292500.3330701}.

\bibitem[Ansel et~al.(2024)Ansel, Yang, He, Gimelshein, Jain, Voznesensky, Bao,
  Bell, Berard, Burovski, Chauhan, Chourdia, Constable, Desmaison, DeVito,
  Ellison, Feng, Gong, Gschwind, Hirsh, Huang, Kalambarkar, Kirsch, Lazos,
  Lezcano, Liang, Liang, Lu, Luk, Maher, Pan, Puhrsch, Reso, Saroufim,
  Siraichi, Suk, Suo, Tillet, Wang, Wang, Wen, Zhang, Zhao, Zhou, Zou, Mathews,
  Chanan, Wu, and Chintala]{Ansel_2024}
Jason Ansel, Edward Yang, Horace He, Natalia Gimelshein, Animesh Jain, Michael
  Voznesensky, Bin Bao, Peter Bell, David Berard, Evgeni Burovski, Geeta
  Chauhan, Anjali Chourdia, Will Constable, Alban Desmaison, Zachary DeVito,
  Elias Ellison, Will Feng, Jiong Gong, Michael Gschwind, Brian Hirsh, Sherlock
  Huang, Kshiteej Kalambarkar, Laurent Kirsch, Michael Lazos, Mario Lezcano,
  Yanbo Liang, Jason Liang, Yinghai Lu, CK~Luk, Bert Maher, Yunjie Pan,
  Christian Puhrsch, Matthias Reso, Mark Saroufim, Marcos~Yukio Siraichi, Helen
  Suk, Michael Suo, Phil Tillet, Eikan Wang, Xiaodong Wang, William Wen,
  Shunting Zhang, Xu~Zhao, Keren Zhou, Richard Zou, Ajit Mathews, Gregory
  Chanan, Peng Wu, and Soumith Chintala.
\newblock {PyTorch 2: Faster Machine Learning Through Dynamic Python Bytecode
  Transformation and Graph Compilation}.
\newblock In \emph{29th ACM International Conference on Architectural Support
  for Programming Languages and Operating Systems, Volume 2 (ASPLOS '24)}. ACM,
  April 2024.
\newblock \doi{10.1145/3620665.3640366}.
\newblock URL \url{https://pytorch.org/assets/pytorch2-2.pdf}.

\bibitem[Kingma and Ba(2017)]{Adam_2017}
Diederik~P. Kingma and Jimmy Ba.
\newblock Adam: {A} {Method} for {Stochastic} {Optimization}, January 2017.
\newblock URL \url{http://arxiv.org/abs/1412.6980}.
\newblock arXiv:1412.6980 [cs].

\bibitem[Falcon and {The PyTorch Lightning
  team}(2019)]{Falcon_PyTorch_Lightning_2019}
William Falcon and {The PyTorch Lightning team}.
\newblock {PyTorch Lightning}, March 2019.
\newblock URL \url{https://github.com/Lightning-AI/lightning}.

\bibitem[Korten et~al.(2024{\natexlab{a}})Korten, Steinbach, and Mirian]{code}
Till Korten, Peter Steinbach, and Najmeh~Sadat Mirian.
\newblock {Virtual Pulse Reconstruction Diagnostic}, November
  2024{\natexlab{a}}.
\newblock URL \url{https://github.com/thawn/VPRD}.

\bibitem[Korten et~al.(2024{\natexlab{b}})Korten, Steinbach, and Mirian]{data}
Till Korten, Peter Steinbach, and Najmeh~Sadat Mirian.
\newblock Training data for "harnessing machine learning for single-shot
  measurement of free electron laser pulse power", November 2024{\natexlab{b}}.
\newblock URL \url{https://doi.org/10.14278/rodare.3253}.

\bibitem[Zhu et~al.(2022)Zhu, Lockmann, Czwalinna, and Schlarb]{Zhu_2022}
J.~Zhu, N.~M. Lockmann, M.~K. Czwalinna, and H.~Schlarb.
\newblock Mixed {Diagnostics} for {Longitudinal} {Properties} of {Electron}
  {Bunches} in a {Free}-{Electron} {Laser}.
\newblock \emph{Frontiers in Physics}, 10, July 2022.
\newblock ISSN 2296-424X.
\newblock \doi{10.3389/fphy.2022.903559}.
\newblock URL
  \url{https://www.frontiersin.org/articles/10.3389/fphy.2022.903559}.
\newblock Publisher: Frontiers.

\bibitem[Wilcoxon(1945)]{Wilcoxon_1945}
Frank Wilcoxon.
\newblock Individual {Comparisons} by {Ranking} {Methods}.
\newblock \emph{Biometrics Bulletin}, 1\penalty0 (6):\penalty0 80--83, 1945.
\newblock ISSN 0099-4987.
\newblock \doi{10.2307/3001968}.
\newblock URL \url{https://www.jstor.org/stable/3001968}.
\newblock Publisher: [International Biometric Society, Wiley].

\bibitem[Bonferroni(1936)]{Bonferroni_1936}
Carlo Bonferroni.
\newblock Teoria statistica delle classi e calcolo delle probabilita.
\newblock \emph{Pubblicazioni del R istituto superiore di scienze economiche e
  commericiali di firenze}, 8:\penalty0 3--62, 1936.
\newblock URL \url{https://cir.nii.ac.jp/crid/1570009749360424576}.

\bibitem[Rossbach et~al.(2019)Rossbach, Schneider, and Wurth]{rossbach2019}
Jörg Rossbach, Jochen~R. Schneider, and Wilfried Wurth.
\newblock 10 years of pioneering {X}-ray science at the {Free}-{Electron}
  {Laser} {FLASH} at {DESY}.
\newblock \emph{Physics Reports}, 808:\penalty0 1--74, May 2019.
\newblock ISSN 0370-1573.
\newblock \doi{10.1016/j.physrep.2019.02.002}.
\newblock URL
  \url{https://www.sciencedirect.com/science/article/pii/S0370157319300663}.

\bibitem[Beye and Klumpp(2020)]{flash2020p}
Martin Beye and Stephan Klumpp, editors.
\newblock \emph{{FLASH}2020+ - upgrade of {FLASH}: conceptual design report}.
\newblock Verlag Deutsches Elektronen-Synchrotron, Hamburg, 2020.
\newblock ISBN 9783945931301.
\newblock \doi{10.3204/PUBDB-2020-00465}.
\newblock URL \url{https://bib-pubdb1.desy.de/record/434950}.

\bibitem[Christie et~al.(2019)Christie, Rönsch-Schulenburg, and Vogt]{POLARIX}
Florian Christie, Juliane Rönsch-Schulenburg, and Mathias Vogt.
\newblock A {PolariX} {TDS} for the {FLASH2} {Beamline}.
\newblock In \emph{{39th International Free Electron Laser Conference}}, pages
  328--331. JACOW Publishing, Geneva, Switzerland, November 2019.
\newblock ISBN 978-3-95450-210-3.
\newblock \doi{10.18429/JACoW-FEL2019-WEP006}.
\newblock URL
  \url{https://accelconf.web.cern.ch/fel2019/doi/JACoW-FEL2019-WEP006.html}.

\bibitem[Marchetti et~al.(2017)Marchetti, Aßmann, Beutner, Bopp, Branlard,
  Braun, Catalán~Lasheras, Christie, Craievich, D'Arcy, Decking, Dorda,
  Grudiev, Herrmann, Hoffmann, Hüning, Krebs, Kube, Lederer, Ludwig, Marutzky,
  Marx, McMonagle, Osterhoff, Pedrozzi, Peperkorn, Pfeiffer, Poblotzki, Prat,
  Reiche, Rönsch-Schulenburg, Rolli, Rothenburg, Schlarb, Scholz, Schreiber,
  Vogt, Wagner, Wilksen, Wittenburg, Wuensch, and Zennaro]{POLARIX2}
Barbara Marchetti, Ralph Aßmann, Bolko Beutner, Markus Bopp, Julien Branlard,
  Hans-Heinrich Braun, Nuria Catalán~Lasheras, Florian Christie, Paolo
  Craievich, Richard D'Arcy, Winfried Decking, Ulrich Dorda, Alexej Grudiev,
  Joerg Herrmann, Matthias Hoffmann, Markus Hüning, Olaf Krebs, Gero Kube,
  Sven Lederer, Frank Ludwig, Frank Marutzky, Daniel Marx, Gerard McMonagle,
  Jens Osterhoff, Marco Pedrozzi, Ingo Peperkorn, Sven Pfeiffer, Frauke
  Poblotzki, Eduard Prat, Sven Reiche, Juliane Rönsch-Schulenburg, Kilian
  Rolli, Jens Rothenburg, Holger Schlarb, Matthias Scholz, Siegfried Schreiber,
  Mathias Vogt, Antonio de~Zubiaurre Wagner, Tim Wilksen, Kay Wittenburg,
  Walter Wuensch, and Riccardo Zennaro.
\newblock X-{Band} {TDS} {Project}.
\newblock In \emph{{8th International Particle Accelerator Conference}}, pages
  184--187. JACOW, Geneva, Switzerland, May 2017.
\newblock ISBN 978-3-95450-182-3.
\newblock \doi{10.18429/JACoW-IPAC2017-MOPAB044}.
\newblock URL
  \url{https://accelconf.web.cern.ch/ipac2017/doi/JACoW-IPAC2017-MOPAB044.html}.

\bibitem[Craievich et~al.(2018)Craievich, Aßmann, Bopp, Braun,
  Catalán~Lasheras, Christie, D'Arcy, Dorda, Foese, Ganter, González~Caminal,
  Grudiev, Hoffmann, Hüning, Jonas, Kleeb, Krebs, Lederer, Libov, Marchetti,
  Marx, McMonagle, Osterhoff, Pedrozzi, Poblotzki, Prat, Reiche, Reukauff,
  Schlarb, Schreiber, Tews, Vogt, Wagner, Wuensch, and Zennaro]{POLARIX3}
Paolo Craievich, Ralph Aßmann, Markus Bopp, Hans-Heinrich Braun, Nuria
  Catalán~Lasheras, Florian Christie, Richard D'Arcy, Ulrich Dorda, Manon
  Foese, Romain Ganter, Pau González~Caminal, Alexej Grudiev, Matthias
  Hoffmann, Markus Hüning, Rolf Jonas, Thomas Kleeb, Olaf Krebs, Sven Lederer,
  Vladyslav Libov, Barbara Marchetti, Daniel Marx, Gerard McMonagle, Jens
  Osterhoff, Marco Pedrozzi, Frauke Poblotzki, Eduard Prat, Sven Reiche,
  Matthias Reukauff, Holger Schlarb, Siegfried Schreiber, Gerd Tews, Mathias
  Vogt, Antonio de~Zubiaurre Wagner, Walter Wuensch, and Riccardo Zennaro.
\newblock Status of the {Polarix}-{TDS} {Project}.
\newblock In \emph{9th {International} {Particle} {Accelerator} {Conference}},
  pages 3808--3811. JACOW Publishing, Geneva, Switzerland, June 2018.
\newblock ISBN 978-3-95450-184-7.
\newblock \doi{10.18429/JACoW-IPAC2018-THPAL068}.
\newblock URL
  \url{https://accelconf.web.cern.ch/ipac2018/doi/JACoW-IPAC2018-THPAL068.html}.

\bibitem[Craievich et~al.(2020)Craievich, Bopp, Braun, Citterio, Fortunati,
  Ganter, Kleeb, Marcellini, Pedrozzi, Prat, Reiche, Rolli, Sieber, Grudiev,
  Millar, Catalan-Lasheras, McMonagle, Pitman, Romano, Szypula, Wuensch,
  Marchetti, Assmann, Christie, Conrad, D'Arcy, Foese, Caminal, Hoffmann,
  Huening, Jonas, Krebs, Lederer, Marx, Osterhoff, Reukauff, Schlarb,
  Schreiber, Tews, Vogt, Wagner, and Wesch]{POLARIX4}
P.~Craievich, M.~Bopp, H.-H. Braun, A.~Citterio, R.~Fortunati, R.~Ganter,
  T.~Kleeb, F.~Marcellini, M.~Pedrozzi, E.~Prat, S.~Reiche, K.~Rolli,
  R.~Sieber, A.~Grudiev, W.~L. Millar, N.~Catalan-Lasheras, G.~McMonagle,
  S.~Pitman, V.~del~Pozo Romano, K.~T. Szypula, W.~Wuensch, B.~Marchetti,
  R.~Assmann, F.~Christie, B.~Conrad, R.~D'Arcy, M.~Foese, P.~Gonzalez Caminal,
  M.~Hoffmann, M.~Huening, R.~Jonas, O.~Krebs, S.~Lederer, D.~Marx,
  J.~Osterhoff, M.~Reukauff, H.~Schlarb, S.~Schreiber, G.~Tews, M.~Vogt,
  A.~de~Z. Wagner, and S.~Wesch.
\newblock Novel $x$-band transverse deflection structure with variable
  polarization.
\newblock \emph{Phys. Rev. Accel. Beams}, 23:\penalty0 112001, Nov 2020.
\newblock \doi{10.1103/PhysRevAccelBeams.23.112001}.
\newblock URL
  \url{https://link.aps.org/doi/10.1103/PhysRevAccelBeams.23.112001}.

\bibitem[Haase et~al.(2020)Haase, Jain, Rigaud, Vorkel, Rajasekhar, Suckert,
  Lambert, Nunez-Iglesias, Poole, Tomancak, and Myers]{haase_2020}
Robert Haase, Akanksha Jain, Stéphane Rigaud, Daniela Vorkel, Pradeep
  Rajasekhar, Theresa Suckert, Talley~J. Lambert, Juan Nunez-Iglesias,
  Daniel~P. Poole, Pavel Tomancak, and Eugene~W. Myers.
\newblock Interactive design of {GPU}-accelerated {Image} {Data} {Flow}
  {Graphs} and cross-platform deployment using multi-lingual code generation,
  November 2020.
\newblock URL
  \url{https://www.biorxiv.org/content/10.1101/2020.11.19.386565v1}.
\newblock Pages: 2020.11.19.386565 Section: New Results.

\bibitem[Rigaud et~al.(2024)Rigaud, Haase, Latechre, Soltwedel, Albert,
  Rajasekhar, and Ross]{rigaud_pyclesperanto_2024}
Stéphane Rigaud, Robert Haase, Cherif Latechre, Johannes Soltwedel, Mavin
  Albert, Pradeep Rajasekhar, and Graham Ross.
\newblock {clEsperanto}/pyclesperanto, September 2024.
\newblock URL \url{https://github.com/clEsperanto/pyclesperanto}.
\newblock original-date: 2022-03-31T11:00:44Z.

\bibitem[Otsu(1979)]{Otsu_1979}
Nobuyuki Otsu.
\newblock A {Threshold} {Selection} {Method} from {Gray}-{Level} {Histograms}.
\newblock \emph{IEEE Transactions on Systems, Man, and Cybernetics}, 9\penalty0
  (1):\penalty0 62--66, January 1979.
\newblock ISSN 2168-2909.
\newblock \doi{10.1109/TSMC.1979.4310076}.
\newblock URL
  \url{https://dspace.tul.cz/server/api/core/bitstreams/36abcc1c-cd72-4569-90ed-607017063124/content}.
\newblock Conference Name: IEEE Transactions on Systems, Man, and Cybernetics.

\end{thebibliography}

%%%%%%%%%%%%%%%%%%%%%%%%%%%%%%%%%%%%%%%%%%%%%%%%%%%%%%%%%%%%

\appendix

\section{Appendix / supplemental material}

\subsection{Data Acquisition}

\begin{figure*}[!hbt]
    \centering
    \includegraphics[width=0.9\textwidth]{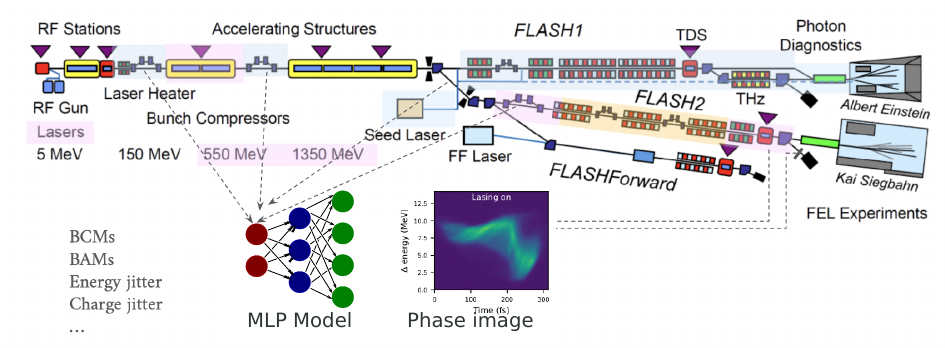}
    \caption{Schematic layout of the FLASH facility at Deutsches Elektronen-Synchrotron (DESY), Germany. Data sources for training the machine learning model are indicated by dashed arrows.}
    \label{fig:FLASH_layout}
\end{figure*}

To collect training data for the machine learning model, we conducted experiments at the Free Electron LASer in Hamburg (FLASH), a state-of-the-art facility at the Deutsches Elektronen-Synchrotron (DESY) in Germany (see Figure \ref{fig:FLASH_layout})\cite{rossbach2019,flash2020p}. FLASH operates by accelerating electron beams using a superconducting linear accelerator (linac), which are subsequently directed through undulator magnets, causing the electrons to follow a sinusoidal path and emit synchrotron radiation. Through self-amplified spontaneous emission (SASE), this radiation is amplified to produce intense, coherent laser pulses. Notably, FLASH comprises two undulator magnet chains—FLASH1 and FLASH2—with the latter known for generating ultra-short pulses, on the order of femtoseconds ($10^{-15}$ seconds).

These ultra-short pulses enable researchers to investigate rapid phenomena, such as chemical reaction dynamics and electron behavior in materials. Users of the FLASH2 beamline often express interest in the FEL pulse profile, which is addressed through the installation of a polarization X-band transverse deflection structure (PolariX TDS) downstream of the FLASH2 undulator line \cite{POLARIX, POLARIX2, POLARIX3, POLARIX4}. 
%By introducing a transverse kick to the electrons, the PolariX TDS deflects the electrons in such a way that the temporal structure of the bunch is mapped onto the spatial domain. This mapping allows for precise measurement of the bunch length and temporal distribution, which are essential for FLASH's performance enhancement. Operating at a specific radiofrequency (RF) of almost 12GHz, the PolariX TDS generates an electromagnetic field that imparts a transverse momentum to the electrons. The RF voltage and phase are meticulously controlled to achieve the desired temporal-to-spatial mapping. The cavity is designed to resolve electron bunch structures on the femtosecond scale, which is critical for characterizing the ultra-short pulses produced by FLASH2.

In the context of our virtual diagnostic tool, the PolariX TDS is essential for data collection. We gather information on the PolariX TDS RF phase, RF voltage, electron beam energy, charge, and the position of the electron beam both before and after the TDS cavity. Additionally, the longitudinal phase space of the electron beam is captured by the YAG screen located after the bending magnet positioned downstream of the TDS cavity (see Figure \ref{fig:FLASH_layout}). \\
We gathered data while the FLASH machine was optimized to deliver FEL radiation at a wavelength of 12 nm for user experiments. The electron beam had a charge of 200 pC and was accelerated to 875 MeV, enabling the FLASH2 beamline to produce 12 nm FEL radiation. During acquisition, the time calibration factor was 1.13 fs/mm, and the energy calibration factor was 21 keV/mm. Our measurements achieved a time resolution of 15.4 fs.

Figure \ref{fig:FLASH_layout} shows the layout of the FLASH machine, highlighting both the FLASH1 and FLASH2 beam lines. The accelerator section includes two bunch compressor sections, BC1 and BC2, each equipped with two bunch compressor monitors (BCMs) and four bunch arrival time monitors (BAMs) as non-invasive diagnostic tools. In total, we recorded $22$ machine parameters (Table \ref{tab:machine_parameters}) and a longitudinal phase space image (Figure \ref{fig:jitter}a) for each electron bunch.

\begin{table}
\centering
\caption{Machine parameters used as model input.}
\begin{tabular}{@{}ll@{}}
\toprule
Parameter name & definition\\ \midrule
BCM.1a & measured data from more sensitive pyroelectric detector after BC1\\
norm. BCM.1a & normalized BCM.1a to the bunch charge \\
BCM.1b& measured data from less sensitive pyroelectric detector in BC1 \\
norm. BCM.1b & normalized BCM.1b to the bunch charge\\
BCM.2a & measured data from more sensitive pyroelectric detector after BC2 \\
norm. BCM.2a & normalized BCM.2a to the bunch charge \\
BCM.2b & measured data from less sensitive pyroelectric detector after BC2 \\
norm. BCM.2b & normalized BCM.2b to the bunch charge \\
BCM.3a & measured data from more sensitive pyroelectric detector in FLASH2 after BC3 \\
norm. BCM.3a& normalized BCM.3a to the bunch charge \\
BCM.3b & measured data from less sensitive pyroelectric detector in FLASH2 after BC3 \\
norm. BCM.3b &normalized BCM.3b to the target of the bunch charge  \\
BAM1-1 & bunch arriving time before BC1  \\
BAM1-2 & bunch exciting time after BC1 \\
BAM2-1 & bunch arriving time before BC2 \\
BAM2-2 & bunch exciting time after BC2 \\
BAM3 &  bunch arriving time before BC3 \\
$\Delta t$ (BAM1-2- BAM1-1) & time delay at BC1 \\ % this is calculated by subtracting BAM2 BC1 from BAM1 BC1
$\Delta t$ (BAM2-2 - BAM2-1) & time delay at BC2  \\ % this is calculated by subtracting BAM2 BC2 from BAM1 BC2 
CHARGE in Gun &  electron bunch charge generated at electron gun  \\
CHARGE in FLASH2 &  Electra bunch charge at FLASH2 beamline \\
ENERGY in FLASH2 & electron beam energy\\
BPM x, y & electron beam position before TDS in x and y directions  \\
\bottomrule
\end{tabular}
\label{tab:machine_parameters}
\end{table}

\subsection{Data preprocessing}
\label{sec:power-calculation}

The electron temporal power profile is calculated from the charge detected in each column of the longitudinal phase space image (Figure \ref{fig:jitter}a) multiplied by the corresponding $\Delta$ energy in $MeV$. The resulting energy weighted charge is projected onto the time axis to calculate the electron temporal power profile (Figure \ref{fig:jitter}b). Image processing was done using the GPU accelerated \texttt{pyclesperanto} \cite{haase_2020, rigaud_pyclesperanto_2024} library. 

\subsubsection{Jittering}

\begin{figure}[htb]
    \centering
    \includegraphics[width=0.7\linewidth]{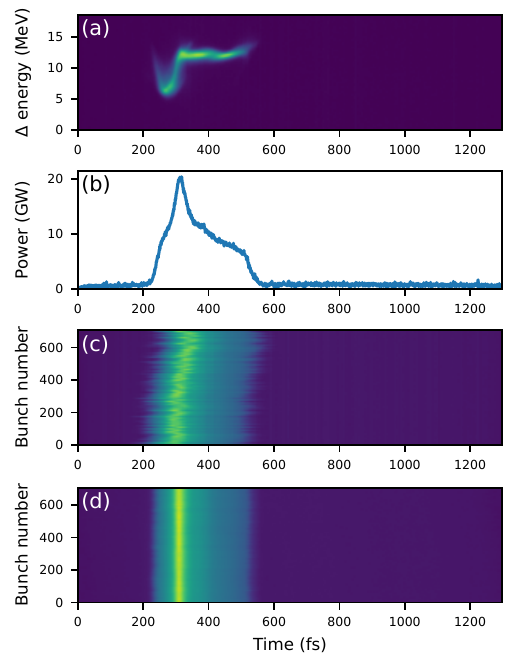}
    \caption{De-jittering. (a) Phase image. (b) Temporal power profile created by weighing the phase image by the energy axis and projecting onto the time axis. (c) Temporal power profiles for 700 samples before de-jittering. (d) Temporal power profiles after de-jittering. }
    \label{fig:jitter}
\end{figure}

To compensate for jitter in the electron bunch arrival time, we calculated the electron temporal power profile for each electron bunch (examples shown in Figure \ref{fig:jitter}a and b). The electron bunches showed considerable temporal jitter (Figure \ref{fig:jitter}c). To compensate for the jitter, the peak power locations were determined from power profiles smoothed by convolving the signal with a gaussian profile with $10$ pixel radius. Then we calculated the offset of each peak to the median peak location and shifted the power profiles accordingly, resulting in excellent alignment (Figure \ref{fig:jitter}d). Finally, we cropped away the parts of the signal that only contained background in every bunch. The location of the signal was determined by segmenting the signal using Otsu's method \cite{Otsu_1979} and cropping to the bounding box of the segmentation with a padding of $10$ pixels. The aligned power profiles were used as labels for the training of the MLP model.

%%%%%%%%%%%%%%%%%%%%%%%%%%%%%%%%%%%%%%%%%%%%%%%%%%%%%%%%%%%%

\end{document}